\documentclass{article}

\usepackage{arxiv}

\usepackage[utf8]{inputenc} 
\usepackage[T1]{fontenc}    
\usepackage{hyperref}       
\usepackage{url}            
\usepackage{booktabs}       
\usepackage{amsfonts}       
\usepackage{nicefrac}       
\usepackage{microtype}      
\usepackage{graphicx}
\usepackage{doi}
\usepackage{multirow}%
\usepackage{amsmath,amssymb,amsfonts}%
\usepackage{amsthm}%
\usepackage{mathrsfs}%
\usepackage[title]{appendix}%
\usepackage{textcomp}%
\usepackage{manyfoot}%
\usepackage{algorithm}%
\usepackage{algorithmicx}%
\usepackage{algpseudocode}%
\usepackage{listings}%
\usepackage{booktabs}
\usepackage[table,xcdraw]{xcolor}
\usepackage{multirow}

\title{Open-Source Periorbital Segmentation Dataset for Ophthalmic Applications}
\author{
    George R.~Nahass \\
    Ophthalmology and Biomedical Engineering \\
    University of Illinois Chicago College of Medicine \\
    Chicago, IL, USA \\
    \texttt{gnahas2@uic.edu} \\
    \And
    Emma~Koehler \\
    Ophthalmology \\
    University of Illinois Chicago College of Medicine \\
    Chicago, IL, USA \\
    \And
    Nicholas~Tomaras \\
    Ophthalmology \\
    University of Illinois Chicago College of Medicine \\
    Chicago, IL, USA \\
    \And
    Danny~Lopez \\
    Ophthalmology \\
    University of Illinois Chicago College of Medicine \\
    Chicago, IL, USA \\
    \And
    Madison~Cheung \\
    Plastic and Reconstructive Surgery \\
    University of Illinois Chicago College of Medicine \\
    Chicago, IL, USA \\
    \And
    Alexander~Palacios \\
    Plastic and Reconstructive Surgery \\
    University of Illinois Chicago College of Medicine \\
    Chicago, IL, USA \\
    \And
    Jeffrey C.~Peterson \\
    Ophthalmology \\
    University of Illinois Chicago College of Medicine \\
    Chicago, IL, USA \\
    \And
    Sasha~Hubschman \\
    Ophthalmology \\
    University of Illinois Chicago College of Medicine \\
    Chicago, IL, USA \\
    \And
    Kelsey~Green \\
    Plastic and Reconstructive Surgery \\
    University of Illinois Chicago College of Medicine \\
    Chicago, IL, USA \\
    \And
    Chad~Purnell \\
    Plastic and Reconstructive Surgery \\
    University of Illinois Chicago College of Medicine \\
    Chicago, IL, USA \\
    \And
    Pete~Setabutr \\
    Ophthalmology \\
    University of Illinois Chicago College of Medicine \\
    Chicago, IL, USA \\
    \And
    Ann Q.~Tran \\
    Ophthalmology \\
    University of Illinois Chicago College of Medicine \\
    Chicago, IL, USA \\
    \texttt{annqtran@uic.edu} \\
    \And
    Darvin~Yi\thanks{Corresponding author} \\
    Ophthalmology and Biomedical Engineering \\
    University of Illinois Chicago College of Medicine \\
    Chicago, IL, USA \\
    \texttt{dyi9@uic.edu} \\
}

\hypersetup{
pdftitle={A template for the arxiv style},
pdfsubject={q-bio.NC, q-bio.QM},
pdfauthor={David S.~Hippocampus, Elias D.~Striatum},
pdfkeywords={First keyword, Second keyword, More},
}

\begin{document}
\maketitle
\begin{abstract}
High quality segmentation of the eyes and lids is an essential step in developing clinically relevant deep learning models for oculoplastic and craniofacial surgery. However, there are currently no publicly available datasets suitable for this purpose. As such, we have developed and validated a novel dataset for oculoplastic segmentation and periorbital distance prediction. Using images from two open-source datasets, we segmented the iris, sclera, lid, caruncle, and brow from cropped eye images. Five trained annotators performed the segmentations, and intergrader reliability was assessed on 100 randomly selected images with a two-week interval, yielding an average Dice score of 0.82 ± 0.01. Intragrader reliability on 20 images averaged a Dice score of 0.81 ± 0.08. To demonstrate the dataset's utility, we trained three DeepLabV3 models following standard procedures. This first-of-its-kind dataset, along with a toolkit for periorbital distance prediction, is publicly available to support the development of clinically useful segmentation models for oculoplastic and craniofacial applications.
\end{abstract}

\section{Introduction}
In ophthalmology, multiple open-source datasets exist for classification and segmentation tasks using optical coherence tomography images, retinal fundus photographs, in vivo confocal microscopy images, and other imaging modalities \cite{porwal_indian_2018, gholami_octid_2019, khan_global_2021, selig_fully_2015}. While the quality and quantity of open-source datasets for images of the inner eye is large, there are far fewer open-source datasets of external eye images for the purposes of training deep learning networks. There are independent datasets for iris segmentation with various labeling of self-reported gender and glasses status, however, these datasets are primarily for use in biometric identification and iris recognition systems \cite{fusek_pupil_2018, proenca_ubirisv2_2010}. 

For a dataset to hold clinical relevance for segmentation tasks in oculoplastic and craniofacial surgery, detailed annotations of meaningful periorbital anatomy need to be created. Current publicly available datasets that have annotations for brows and lids were designed to train algorithms for face recognition and image generation using generative adversarial networks, and, out of the box, lack the appropriate detail required to obtain anatomically accurate segmentation at precise levels \cite{lee_maskgan_2020}.

In the literature, many papers have recently used segmentation of external ocular anatomy as an intermediate step in predicting periorbital distances \cite{van_brummen_periorbitai_2021, shao_deep_2023, shao_automatic_2024}. Clinically, measuring periorbital distances is an important step to track disease progression and monitor treatment efficacy. However manual measurements is a time consuming and error prone process \cite{van_brummen_periorbitai_2021, boboridis_repeatability_2001}. While automatic prediction of periorbital distances via deep learning stands to significantly reduce the time burden of oculoplastic and craniofacial surgeons in the clinic, it also presents an attractive strategy to objectively measure periorbital distances to a high degree of accuracy. For research in this area to move forward, the development of open-source datasets with periorbital annotations created at the level of detail required for prediction of sub millimeter distances is required. As such, we have curated two open-source datasets of external eye images, created annotations of the caruncle, iris, lids, brows, and scleras, and demonstrated their efficacy in training deep learning models with them. Additionally, we have open-sourced a toolkit to predict periorbital distances from segmentation masks and ground truth measurements generated from our dataset to encourage future research and model development in this space.

\section{Methods}
\subsection{Data Acquisition}
	
Open-source data was used as the foundation of the dataset. Images were sourced from the Chicago Facial dataset (CFD) and the CelebAMask-HQ dataset (Celeb) \cite{lee_maskgan_2020, ma_chicago_2015, ma_chicago_2021, lakshmi_india_2021}. All images were cropped to include the eyes and periorbital regions using Mediapipe Facemesh \cite{kartynnik_real-time_2019}. 827 images were included from the CFD, and 2015 images were included from Celeb.

\subsection{Annotation in CVAT}
	
\begin{figure}[t]
    \centering
    \includegraphics[width=\linewidth]{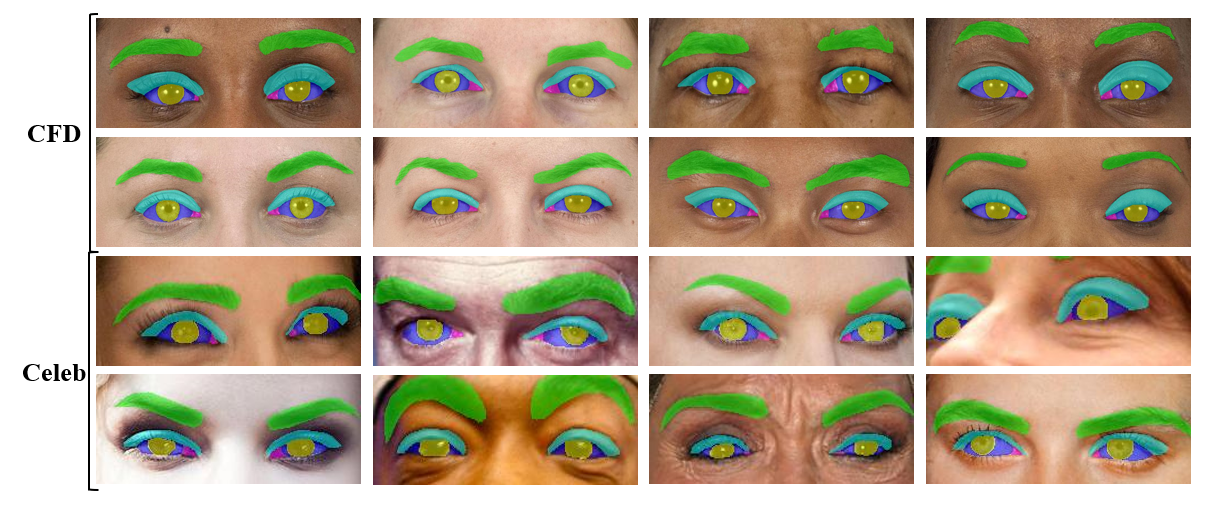}
    \caption{Representative images and annotations from the CFD and Celeb dataset used to construct the dataset described here.}
    \label{fig:eye_grid}
\end{figure}

Five annotators were trained to use the Computer Vision Annotation Tool (CVAT) to draw masks denoting the iris, sclera, lids, caruncles, and brows. Existing brow and sclera annotations were used as starter masks for the Celeb dataset. For lid annotations, the lid crease was used as the superior margin and the superior rim of the sclera was used as the inferior margin. The lid annotation was bounded by the shortest line between the medial or lateral canthus to the lid crease, or the termination of the lid crease against the superior scleral margin. Upper eyelashes were included in the lid annotation, and epicanthal folds were not annotated. Iris annotations were bounded inferiorly and superiorly by the scleral margin. Scleral annotations were made from the lateral canthus to the caruncle, and the caruncle was annotated only if present. Annotations were exported from CVAT in COCO format and post-processed so every anatomical structure is denoted by a specific pixel value ranging 1-5. Representative images of eyes with full annotations from both CFD and the Celeb dataset can be found in Figure \ref{fig:eye_grid}.

\subsection{Intra and Inter Grader Validation}
Each annotator annotated an additional 100 images (60 from Celeb, 40 from CFD), where 20 images were ones they had annotated in the past. A minimum of two weeks had elapsed from the time of initial annotation to permit fair intra-grader evaluation. For the 100 images, the Dice score was computed pairwise between each annotator according to Equation \ref{dice}.

\begin{equation}
    \label{dice}
    Dice = \frac{2(\mathbf{X}\cap{\mathbf{Y}})}{|\mathbf{X}| + |\mathbf{Y}|}
\end{equation}

\subsection{Segmentation Pipeline}
We trained segmentation models on both the CFD and Celeb datasets. A DeepLabV3 segmentation network with a ResNet-101 backbone pretrained on ImageNet1K was implemented from Torchvision \cite{maintainers_torchvision_2016}. The final layer was modified to output six output classes, and the model was trained for $500$ steps.  A train test split of $80/20$ was used with cross-entropy loss and a batch size of 16. Adam optimization was used with a learning rate of .0001 and beta values of $.9$ and $.99$. 	

\begin{figure}[h]
    \centering
    \includegraphics[width=0.9\linewidth]{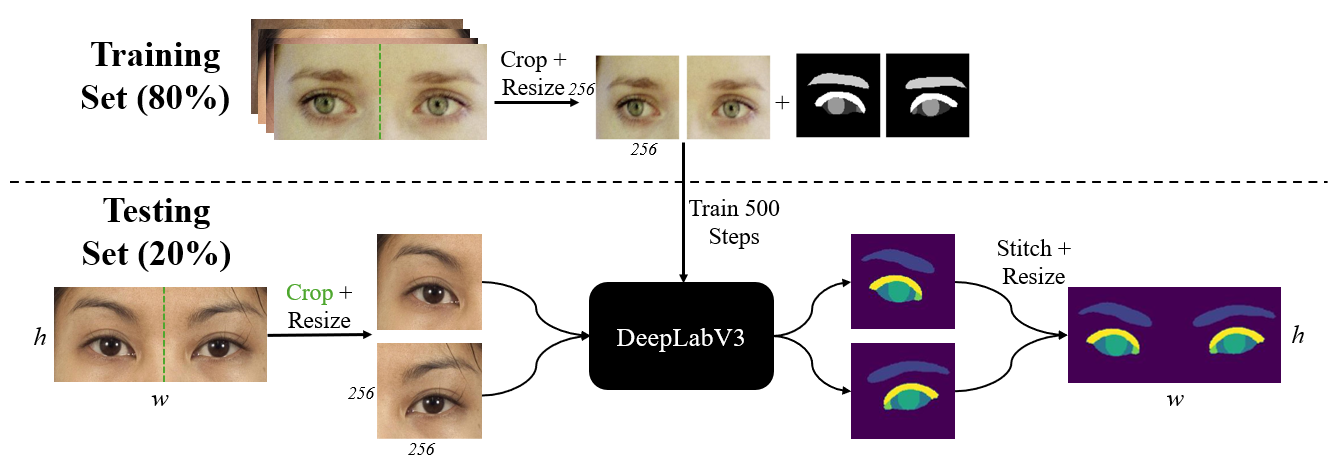}
    \caption{Schematic of preprocessing and training pipeline. Full details are described in the methods, but briefly, the dataset was split using an 80/20 train test split. The input image was split at the midline, and both halves of the image (and label) were resized to 256x256. A DeepLabV3 model with a ResNet101 backbone pretrained on ImageNet1K was trained for 500 steps. The same preprocessing procedure was used at test time. Following segmentation, the left and right halves of the image were resized and stitched back together such that the full segmentation mask was the same size as the input.}
    \label{fig:pipeline}
\end{figure}

Prior to training and prediction, images were split at the midline and resized to $256x256$. At test time, the same process was applied, and the resulting segmentation maps of both halves of the image were recombined using the same aspect ratio as the initial image. Dice score on the test set was computed using the recombined image and the original segmentation mask according to Equation \ref{dice}. The entire segmentation pipeline can be seen in Figure \ref{fig:pipeline}.

\subsection{Prediction of Periorbital Distances}

Periorbital distances were generated using the human annotated segmentation masks to provide a benchmark set of results. The iris diameter was set to a scale of 11.71 millimeters (mm), which was used to derive pixel to mm conversions as described by Van Brummen et.al.\cite{van_brummen_periorbitai_2021}. Scleral show was calculated only if present using the scleral and iris margins. Margin to Reflex Distance 1 and 2 (MRD 1 and 2) were calculated as the distance from the center of the iris to the superior or inferior lid.

Inner and outer canthal distance (ICD, OCD) and interpupillary distance (IPD) were calculated as the distance between the medial canthus, lateral canthus, and iris center, respectively. Brow heights were obtained at the medial canthus, lateral canthus, and iris center. Canthal tilt, canthal height, and vertical dystopia were also measured using the medial canthus, lateral canthus, and iris center landmarks. Vertical palpebral fissure was computed as the sum of MRD 1 and 2 and the horizontal palpebral fissure was calculated using the x coordinates of the medial and lateral canthus. 

\subsection{Hardware}
All batch experiments were performed on three 1080TI Nvidia GPUs. All code was written in Python 3.8 and standard machine learning packages were used for all training purposes \cite{paszke_pytorch_2019, harris_array_2020, pedregosa_scikit-learn_2011}.

\section{Results}\label{results}

\subsection{Description of the Dataset}

Both the CFD and Celeb datasets are open source, so the only available metadata is that which was originally published with the dataset. A subset of the original Celeb dataset was randomly chosen for annotation. There are 827 and 2015 images and annotations for the CFD and Celeb dataset respectively. Within the CFD dataset, the counts of self-reported races are 109 Asian, 197 Black, 142 Indian, 108 Latin, 88 Mixed, and 183 White.  Of the 827 images, 421 identified as female and 406 identified as male \cite{ma_chicago_2015, ma_chicago_2021, lakshmi_india_2021}. The Celeb dataset did not include any racial or gender metadata.

\begin{figure}[h]
    \centering
    \includegraphics[width=\linewidth]{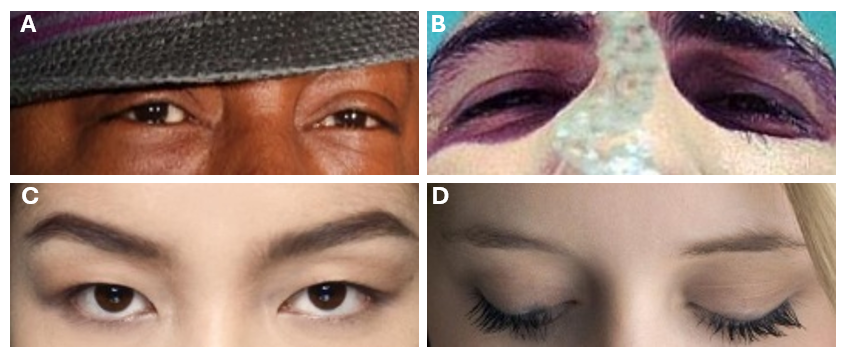}
    \caption{Examples of images lacking certain annotations. A) Image lacking brow annotations, B) Image lacking caruncle annotations, C) Image lid annotations D) Image lacking sclera and iris annotations}
    \label{fig:no_annot}
\end{figure}

For both the CFD and Celeb datasets, the total number of anatomical objects annotated across all images can be found in Table \ref{tab:metadata}. Not all images had visible caruncles, and epicanthal lids were not annotated. An example of how different types of lids were annotated is shown in Figure \ref{fig:lid_var}, and examples of images lacking certain annotations can be found in Figure \ref{fig:no_annot}.

\begin{table}[ht]
\centering
\resizebox{.8\columnwidth}{!}{%
\begin{tabular}{@{}|
>{\columncolor[HTML]{FFFFFF}}c |
>{\columncolor[HTML]{FFFFFF}}c |
>{\columncolor[HTML]{FFFFFF}}c |
>{\columncolor[HTML]{FFFFFF}}c |
>{\columncolor[HTML]{FFFFFF}}c |
>{\columncolor[HTML]{FFFFFF}}c |
>{\columncolor[HTML]{FFFFFF}}c |
>{\columncolor[HTML]{FFFFFF}}c |@{}}
\toprule
\textbf{Dataset} & Total   Images & \%   Full Annotation & Sclera & Iris & Caruncle & Brow & Lid  \\ \midrule
CFD              & 827            & 0.97                 & 827    & 827  & 827      & 827  & 803  \\ \midrule
Celeb            & 2015           & 0.89                 & 2002   & 2002 & 1884     & 1994 & 1892 \\ \bottomrule
\end{tabular}%
}
\caption{Metadata of our dataset from the two constituent open-source datasets showing the total number of images, the percent of images that have every class present, and the number of images containing each class.}
\label{tab:metadata}
\end{table}

\subsection{Segmentation Results}
\begin{figure}
    \centering
    \includegraphics[width=\linewidth]{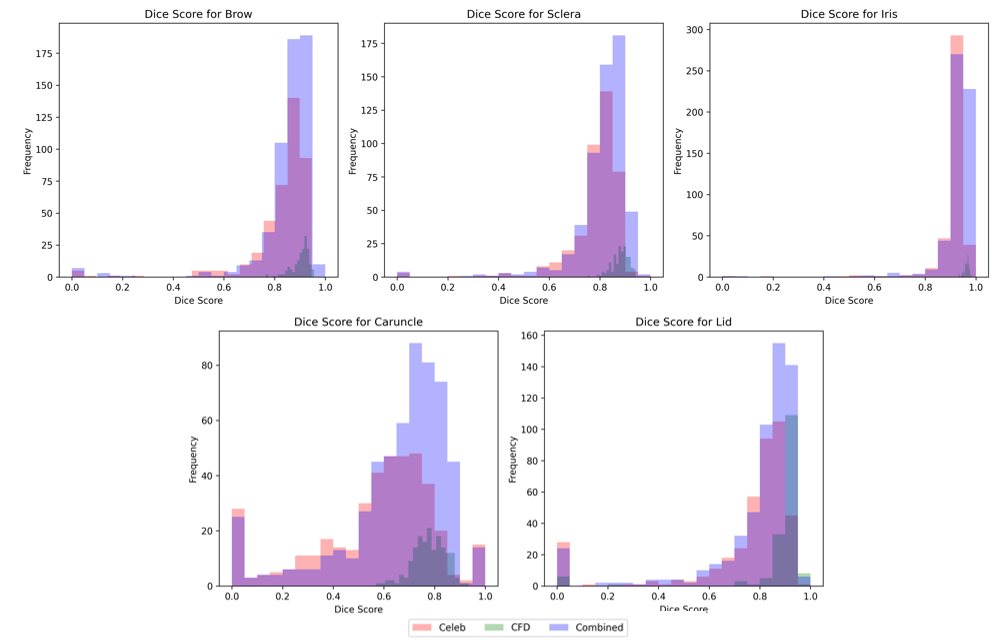}
    \caption{Histograms of the Dice score for each class from each model trained.}
    \label{fig:histos}
\end{figure}

We trained three deep learning networks for segmentation using 1) the CFD, 2) the Celeb dataset, and 3) both datasets combined. In all cases, segmentation of the iris, sclera, and brow was robust, with average Dice scores ranging from $.78$ to $.96$. On the Celeb dataset, caruncle and lid segmentation were less robust on average, with Dice scores being $.56$ and $.75$ respectively. Caruncle and lid segmentation on the CFD was improved compared to the Celeb dataset, with average dice scores of $.78$ and $.87$, respectively. Dice scores of a segmentation network trained and tested on both datasets were between those from the models trained on only the CFD or Celeb dataset. The Dice scores for all experiments can be found in Table \ref{tab:seg_dice}, and histograms of all dice scores can be found in Figure \ref{fig:histos}.

\begin{table}[ht]
\centering
\resizebox{.8\columnwidth}{!}{%
\begin{tabular}{@{}|c|c|c|c|c|c|@{}}
\toprule
\cellcolor[HTML]{FFFFFF}\textbf{} & \textbf{Sclera} & \textbf{Iris} & \textbf{Brow} & \textbf{Caruncle} & \textbf{Lid} \\ \midrule
\textbf{Celeb}    & 0.78  ± 0.11 & 0.91  ± 0.08 & 0.83  ± 0.14 & 0.57  ± 0.24 & 0.75  ± 0.23 \\ \midrule
\textbf{CFD}      & 0.88  ± 0.03 & 0.96  ± 0.01 & 0.90  ± 0.03 & 0.78  ± 0.07 & 0.87  ± 0.18 \\ \midrule
\textbf{Combined} & 0.81  ± 0.11 & 0.93  ± 0.08 & 0.85  ± 0.14 & 0.65  ± 0.22 & 0.79  ± 0.21 \\ \bottomrule
\end{tabular}%
}
\caption{Average Dice scores for each class. Celeb denotes a model trained only on the Celeb annotations, CFD denotes a model trained only on the CFD annotations, and Combined denotes a model trained when the CFD and Celeb dataset was combined.}
\label{tab:seg_dice}
\end{table}

\subsection{Intra and Intergrader Variation}
\begin{figure}[ht]
    \centering
    \includegraphics[width=.8\linewidth]{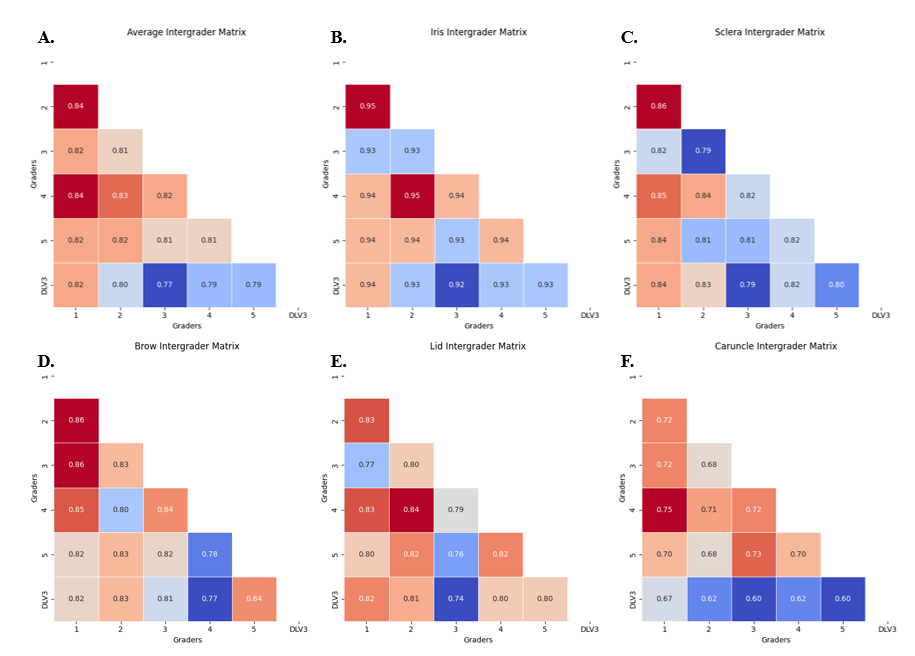}
    \caption{Pairwise matrices representing intergrader agreement as the average Dice score between graders or DeepLabV3 over 100 randomly sampled images. A) The average pairwise Dice score between all graders, and B-F) represent the Dice score on the iris, sclera, brow, lid, and caruncle classes respectively.}
    \label{fig:intergrader}
\end{figure}

To evaluate the quality of our dataset, we randomly sampled 100 images and performed intergrader evaluation. When the same five annotators annotated the same 100 images, the average Dice Score between pairwise graders was $.82\pm.01$. The average dice score for iris, sclera, brow, lid, and caruncle annotations across all annotators was $.94\pm.01$, $.83\pm.02$, $.83\pm.02$, $.81\pm.03$, $.71\pm .02$. Pairwise matrices showing the Dice scores between all graders and a DeepLabV3 segmentation network can be seen in Figure \ref{fig:intergrader}. As the intergrader sample set consisted of both CFD and Celeb Dataset images, the DeepLabV3 model trained only on the individual dataset was used for each image. For example, all CFD images in the intergrader set were segmented using the DeepLabV3 model trained only on the CFD dataset. The agreement with the DeepLabV3 network between all graders was generally lower than agreement between graders, but not significantly.

We also asked each of our five annotators to annotate the same 20 images two times after at least a 2-week forgetting period. On average, the same grader annotated the same image very reproducibly with an average Dice Score of $.81\pm.08$ across all annotations (Table \ref{tab:intra}).  Across all graders, the iris was the most consistently reproduced with an average intragrader Dice score of $.94$. Sclera and lid annotations were reproduced with an average Dice score of $.82$ and $.83$ across all graders, while brow and caruncle were the most challenging anatomical part to reproduce having an average intragrader Dice score of $.77$ and $.71$ respectively.

\begin{table}[b!]
\centering
\resizebox{.5\columnwidth}{!}{%
\begin{tabular}{@{}|
>{\columncolor[HTML]{FFFFFF}}c |
>{\columncolor[HTML]{FFFFFF}}c |
>{\columncolor[HTML]{FFFFFF}}c |
>{\columncolor[HTML]{FFFFFF}}c |
>{\columncolor[HTML]{FFFFFF}}c |
>{\columncolor[HTML]{FFFFFF}}c |@{}}
\toprule
\textbf{}           & \textbf{Brow} & \textbf{Sclera} & \textbf{Iris} & \textbf{Caruncle} & \textbf{Lid} \\ \midrule
\textbf{Grader   1} & 0.66          & 0.84            & 0.95          & 0.75              & 0.85         \\ \midrule
\textbf{Grader   2} & 0.72          & 0.80            & 0.94          & 0.66              & 0.83         \\ \midrule
\textbf{Grader   3} & 0.80          & 0.78            & 0.92          & 0.65              & 0.72         \\ \midrule
\textbf{Grader   4} & 0.89          & 0.88            & 0.96          & 0.77              & 0.90         \\ \midrule
\textbf{Grader   5} & 0.88          & 0.82            & 0.96          & 0.80              & 0.88         \\ \midrule
\textbf{Average}    & 0.77          & 0.82            & 0.94          & 0.71              & 0.83         \\ \bottomrule
\end{tabular}%
}
\caption{Intragrader agreement represented as Dice score. Each grader annotated the same 20 images after a two-week forgetting period, and the Dice score is shown for each class. Average represents the average intragrader agreement for each class.}
\label{tab:intra}
\end{table}
\section{Discussion}\label{disc}

In the literature, periorbital segmentation is a primary intermediate step for periorbital distance prediction \cite{van_brummen_periorbitai_2021, rana_artificial_2024}. Periorbital distances are highly time-consuming to obtain clinically, yet they provide valuable information in the clinical decision-making process with respect to disease tracking and treatment monitoring \cite{shao_deep_2023, chen_smartphone-based_2021}. Prior studies using segmentation as an intermediate step for periorbital distance prediction currently rely on splitting relatively rare clinical data into train and test sets. This requires using a large amount of a precious resource for training purposes and limits the size of the test set, making it challenging to assess the generalizability of such models \cite{alauthman_enhancing_2023}. Furthermore, it has been shown that training open-source datasets can lead to accurate segmentation on images of craniofacial and oculoplastic pathology collected both in the wild and using professional photographers \cite{nahass2024stateoftheartperiorbitaldistanceprediction}. However, there is currently an absence of open-source datasets designed specifically for periorbital segmentation.

\begin{figure}[h]
    \centering
    \includegraphics[width=\linewidth]{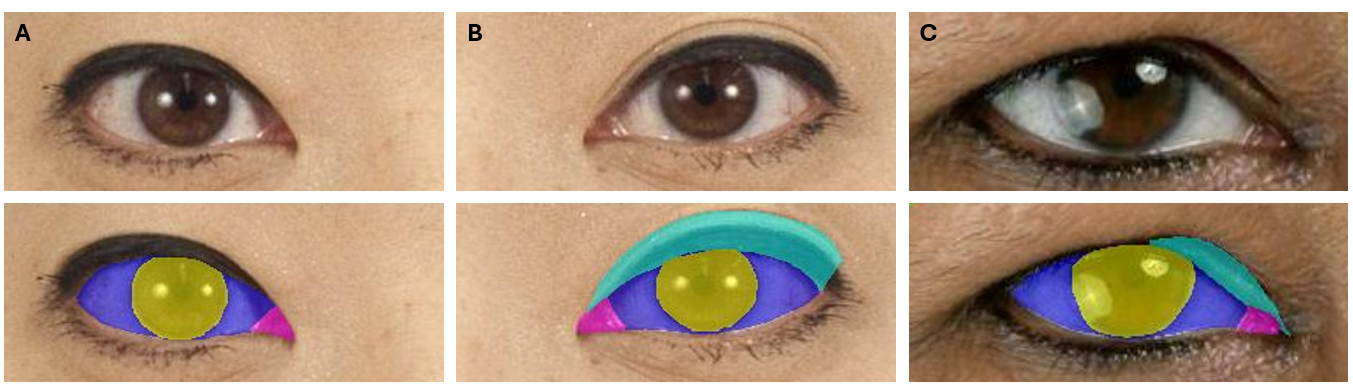}
    \caption{Examples of how different types of lids were annotated using the lid crease as a guide. A) No visible lid crease, so no annotation was created. B) When the lid crease was present, the lid annotation included any visible lashes so the boundaries of the masks align with the sclera mask C) In the event of a partially visible lid crease, only regions of the lid inferior to the lid crease were annotated.}
    \label{fig:lid_var}
\end{figure}

In existing datasets, the lack of caruncle annotation often leads to the medial canthus not being well defined and lid margins that are not adequately demarcated by existing annotations. To solve this, we have leveraged existing open-source datasets to develop a comprehensive dataset designed for medical grade segmentation of relevant periorbital anatomy \cite{lee_maskgan_2020, ma_chicago_2015, ma_chicago_2021, lakshmi_india_2021}. On the CFD and Celeb dataset we have created iris, lid, sclera, brow, and caruncle annotations in accordance with standard accepted definitions of this anatomy and through consultation with oculoplastic surgeons at our institution. We only annotated the object if it was present in the image. For example, if the eye was closed, no sclera or iris would be annotated. We also only annotated lids and portions of lids that were defined by a clear lid crease. As such, epicanthal lids were not annotated. An example of the criteria used for lid annotation can be seen in Figure \ref{fig:lid_var}. While the CFD consists of images that are directly facing forward, the Celeb dataset contains images of subjects at various angles and lighting conditions. For this reason, incorporating the Celeb dataset into training strategies may be a valuable approach for training models with the utility on in-the-wild clinical images as the distribution of images from this dataset is broader. 
To demonstrate the efficacy of these annotations for training segmentation models, we trained networks on both the CFD and Celeb datasets. We achieved robust segmentation on all the anatomical objects of interest.  In our prior work, we have shown that accurate sclera and brow segmentation can be achieved on images displaying pathology through training on as few as 1000 open-source images. We replicated this finding using our dataset \cite{nahass2024stateoftheartperiorbitaldistanceprediction}. The models we trained here did have lower performance on lids and caruncles relative to the iris, sclera, and brows. As such, we plan to expand this dataset to include more annotations for lids and caruncles and have provided all the relevant CVAT documentation to allow for community engagement in growing the dataset.

As our dataset was built on originally open-source datasets, we only include metadata available in the original publications. Of the two datasets we compiled, only the CFD dataset included metadata such as the self-reported racial and gender identity of the patients. The CFD component of our dataset is split $.51$ to $.49$ percent female to male, and there are a range of racial identities represented. This metadata may be useful in evaluating any potential bias of future segmentation models trained for periorbital segmentation.

We have validated the annotations through intra and intergrader comparisons and showed that our annotations are both high quality and reproducible.  However, there is some inherent variability when the different graders annotate the same image and when the same annotator annotates the same image twice. This effect was most pronounced for the caruncle, brow, and lid annotations. Even though the graders were well trained, there is still inherent subjectivity in demarcating these anatomical structures on close-up images, particularly when the images are lower resolution (as is the case with the Celeb dataset) and the structures are relatively small compared to the entire image. Additionally, the agreement with the DeepLabV3 model with all graders was the lowest on average, indicating that while the deep learning models perform relatively well, there is still room for significant improvement in model development. On intragrader analysis, variation between two annotation sessions is most pronounced with the brow and caruncle classes, likely due to the nature of these classes.  

\begin{figure}
    \centering
    \includegraphics[width=0.75\linewidth]{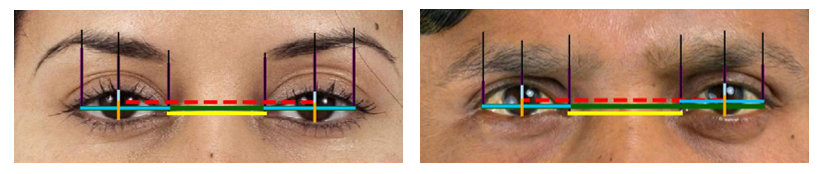}
    \caption{Periorbital distances on two images from the CFD dataset. These distances can be calculated using the toolkit, which we have made available via API, and the periorbital distances from the CFD dataset have been released as a benchmark dataset.}
    \label{fig:distances}
\end{figure}

In addition to open-sourcing our datasets, we have also created a toolkit for distance prediction from periorbital segmentation masks. This code has been released via an application programming interface (API) so others in the community can use it as a tool in future work related to the prediction of periorbital distances. Within the API, our models can be accessed, and full facial images or cropped images can be submitted for segmentation with or without periorbital distance calculation from the masks. Additionally, users can submit segmentation masks from their own models to obtain the periorbital distances using our analysis pipeline. The Python package can be found \href{https://pypi.org/project/periorbital-package/}{here}. An example of an image with periorbital distances denoted by the toolkit can be seen in Figure \ref{fig:distances}. To provide a benchmark for iterative improvement, we have released the periorbital distances of the ground truth segmentation masks for the CFD images in our dataset as these images are of individuals facing forward in a standard position. We hope that the community will find utility in these annotations, the toolkit, and the benchmark distances for future research in this space.

\section{Dataset and Code Availability}

The entire dataset is available to download \href{https://zenodo.org/records/13916845}{from Zenodo}. Code used to train all models can be found at \href{https://github.com/aiolab/periorbital-dataset}{the Artificial Intelligence in Ophthalmology center's Github.}

\bibliographystyle{unsrt}
\bibliography{main}

\section{Supplemental Figures}\label{figs}

\end{document}